\documentclass[journal]{IEEEtran}
\usepackage[T1]{fontenc}
\usepackage{graphicx}
\usepackage{times}
\usepackage{helvet}
\usepackage{courier}
\usepackage{amsmath}
\usepackage{algorithm}
\usepackage{algorithmic}
\usepackage{csquotes} 
\usepackage{color}
\usepackage{paralist}
\usepackage{amssymb}
\usepackage{indentfirst}
\usepackage{subfigure}
\usepackage{float}
\usepackage{multirow}
\usepackage{cite}
\usepackage{mathrsfs}

\usepackage{mathrsfs} 
\usepackage{amsfonts}
\usepackage{todonotes}
\usepackage{pgfplots} 
\usepackage{epstopdf}
\usepackage{epsfig}
\pgfplotsset{compat=newest}
\usepackage{color}
\definecolor{forestgreen}{RGB}{0,139,69}
\usepackage{enumitem}
\usepackage{xcolor}
\definecolor{citecolor}{HTML}{0071bc}
\usepackage[colorlinks, linkcolor=red,  anchorcolor=blue, citecolor=citecolor]{hyperref} 

\usepackage{xcolor}
\definecolor{SeaGreen4}{RGB}{0,205,102} 
\definecolor{SlateBlue}{RGB}{106,90,205} 
\definecolor{DarkRed}{RGB}{178,34,34} 
	
\usepackage[switch]{lineno}
\nolinenumbers

\usepackage{textcomp,booktabs}
\usepackage{amssymb}
\usepackage{pifont}
\newcommand{\cmark}{\ding{51}}%

\usepackage{makecell}

\usepackage{colortbl}
\definecolor{mygray}{gray}{.9}
\definecolor{mypink}{rgb}{.99,.91,.95}
\definecolor{mycyan}{cmyk}{.3,0,0,0}

\begin{document}

\title{Segment Any Vehicle: Semantic and Visual Context Driven SAM and A Benchmark}

\author{ Xiao Wang, \emph{Member, IEEE}, Ziwen Wang, Wentao Wu, Anjie Wang, Jiashu Wu, Yantao Pan, Chenglong Li*

\thanks{
$\bullet$ Xiao Wang, Ziwen Wang, Anjie Wang, and Jiashu Wu are with the School of Computer Science and Technology, Anhui University, Hefei 230601, China. 
(email: e24201001@stu.ahu.edu.cn, xiaowang@ahu.edu.cn) 

$\bullet$ Yantao Pan is with Chery, Wuhu, Anhui Province, China (email: pytmail@126.com)

$\bullet$ Wentao Wu, Chenglong Li are with the School of Artificial Intelligence, Anhui University, Hefei 230601, China. 
(email: wuwentao0708@163.com, lcl1314@foxmail.com)
} 
\thanks{* Corresponding Author: Chenglong Li (lcl1314@foxmail.com)}   
}

\markboth{ IEEE Transactions on ***, 2025 } 
{Shell \MakeLowercase{\textit{et al.}}: Bare Demo of IEEEtran.cls for IEEE Journals}

\maketitle

\begin{abstract}

With the rapid advancement of autonomous driving, vehicle perception, particularly detection and segmentation, has placed increasingly higher demands on algorithmic performance. Pre-trained large segmentation models, especially Segment Anything Model (SAM), have sparked significant interest and inspired new research directions in artificial intelligence. However, SAM cannot be directly applied to the fine-grained task of vehicle part segmentation, as its text-prompted segmentation functionality is not publicly accessible, and the mask regions generated by its default mode lack semantic labels, limiting its utility in structured, category-specific segmentation tasks. To address these limitations, we propose SAV, a novel framework comprising three core components: a SAM-based encoder-decoder, a vehicle part knowledge graph, and a context sample retrieval encoding module. The knowledge graph explicitly models the spatial and geometric relationships among vehicle parts through a structured ontology, effectively encoding prior structural knowledge. Meanwhile, the context retrieval module enhances segmentation by identifying and leveraging visually similar vehicle instances from training data, providing rich contextual priors for improved generalization. Furthermore, we introduce a new large-scale benchmark dataset for vehicle part segmentation, named VehicleSeg10K, which contains 11,665 high-quality pixel-level annotations across diverse scenes and viewpoints. We conduct comprehensive experiments on this dataset and two other datasets, benchmarking multiple representative baselines to establish a solid foundation for future research and comparison. 
Both the dataset and source code of this paper will be released on \url{https://github.com/Event-AHU/SAV} 
\end{abstract}

\begin{IEEEkeywords}
Semantic Segmentation; Knowledge Graph; Context Sample Augmentation; Vehicle Perception; Segment Anything Model  
\end{IEEEkeywords}

\IEEEpeerreviewmaketitle

\section{Introduction}

\IEEEPARstart{V}{ehicle} segmentation~\cite{elhassan2024real} plays a crucial role in modern intelligent transportation systems~\cite{wu2024vfmDeT, wang2024vehicleMAE, wang2022PARSurvey}, serving as a fundamental component for autonomous driving, advanced driver-assistance systems (ADAS), and intelligent traffic management. The precise localization and identification of vehicle parts also facilitate advanced functionalities, including damage assessment, automated parking assistance, and detailed vehicle analysis for insurance and maintenance purposes. However, accurately segmenting vehicle components at the part level presents unique challenges due to the complex structural relationships between parts, varying lighting conditions, and diverse vehicle appearances across different models and viewpoints.

\begin{figure*}[t]
\centering
\includegraphics[width=1\linewidth]{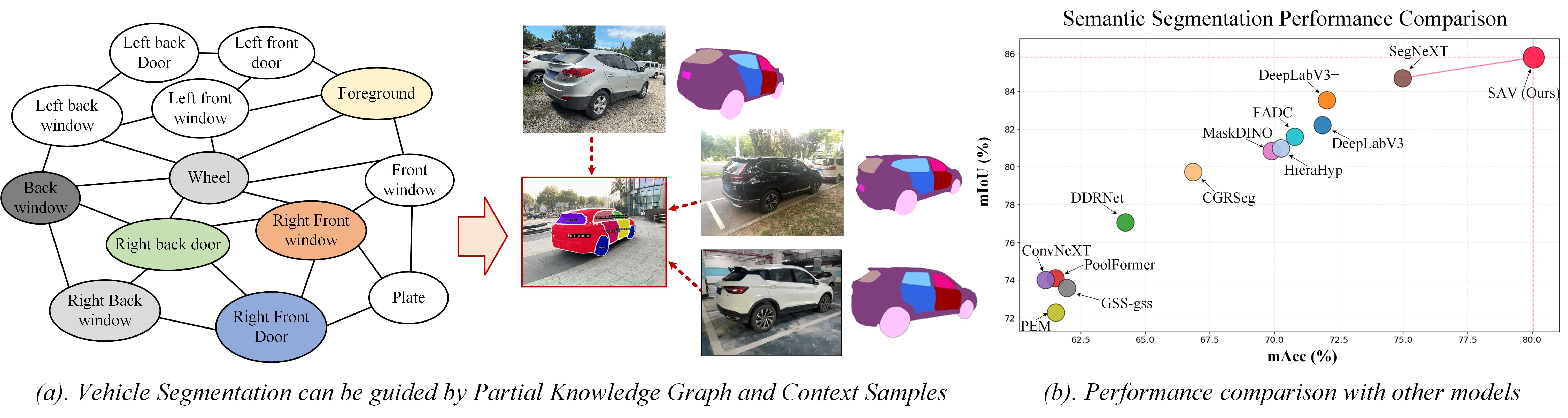}
\caption{ 
(a) Vehicle segmentation guided by partial knowledge graph and context samples. The knowledge graph captures structural relationships between vehicle parts, while context samples provide visual appearance guidance from similar vehicles. 
(b) Comparison between our model with state-of-the-art methods on semantic segmentation.} 
\label{fig:firstIMG}
\end{figure*}

In the era of deep learning, early segmentation models (e.g., DeepLab~\cite{chen2017deeplab}, FCN~\cite{long2015fully}, UNet~\cite{ronneberger2015u}) generally adopted the encoder-decoder framework, achieving good segmentation accuracy with the help of CNN models like ResNet~\cite{he2016deep}. However, limited by the amount of human-annotated segmentation data, these segmentation models still could not be used in challenging scenarios. Recent advances in artificial intelligence have witnessed the emergence of foundation models that demonstrate remarkable capabilities in language understanding and generation~\cite{devlin2019bert}, and multi-modal aggregation~\cite{cui2024survey}. Additionally, the Segment Anything Model (SAM)~\cite{kirillov2023segment}, trained on over one billion masks, has revolutionized image segmentation through its unprecedented zero-shot transfer capabilities and high-quality mask generation.

Building upon SAM's success, some researchers have attempted to extend its capabilities for specific applications. 
For example, Semantic-SAM~\cite{li2307semantic} incorporates CLIP embeddings to enable text-prompted segmentation. PerSAM~\cite{zhang2023personalize} personalizes the model through reference examples. 
HQ-SAM~\cite{ke2023segment} focuses on improving segmentation quality through supplementary processing modules. 
In parallel, significant progress has been made in vehicle perception and segmentation for autonomous driving applications~\cite{zhao2024autonomous}. Recent advances include multi-scale adaptive attention mechanisms for driving scene segmentation~\cite{liu2023semantic} and comprehensive reviews of real-time semantic segmentation approaches that leverage CNNs, Transformers, and hybrid architectures for autonomous vehicle perception~\cite{elhassan2024real}.

Despite the significant progress, these works are still limited by the following issues: 
1). Although SAM supports text-prompted segmentation, the functionality is not publicly accessible via its official interface. SAM can produce high-quality segmentation masks, but it lacks semantic labeling, which limits its direct applicability to fine-grained tasks such as vehicle part segmentation. 
2). Existing segmentation models mainly focus on local visual features, yet fail to explicitly model the strong spatial and structural relationships among vehicle parts, i.e., the prior knowledge that is crucial for anatomically consistent parsing. 
3). Contextual information from surrounding regions or related samples can provide valuable priors for feature learning, but remains underexploited in current vehicle segmentation approaches. 
Therefore, it is natural to raise the following question: \textit{How can we effectively incorporate contextual cues and structural priors into vehicle part segmentation to achieve accurate, semantically meaningful segment results?}

In this paper, we propose a novel semantic and visual context driven segment anything vehicle model, termed SAV. The key insight of our framework lies in effectively capturing the spatial and geometric relationships among vehicle parts through the construction of a vehicle part ontology, and enhancing current vehicle segmentation by retrieving visual context, i.e., the images of vehicles from similar viewpoints, to improve segmentation performance. As shown in Fig.~\ref{fig:firstIMG}, given the vehicle image, we first employ the SAM encoder to obtain its multi-scale feature representations. Meanwhile, we treat the vehicle partial labels as the entity of the knowledge graph and build their relations based on the vehicle keypoints (analogous to human skeletal joints, where only adjacent keypoints are connected by edges) and the co-occurrence frequency among different labels. Then, we adopt the GATv2~\cite{brody2021attentive} to encode the vehicle graph and fuse with the vision features using CDT and PEEM operations. Then, we feed the raw vision features and VPKG-enhanced representations into the SAM decoder to output the RoI features. The vehicle context samples are retrievaled and injected into the decoding phase to further augment the representations. Finally, we adopt the cross-attention and logit operation to get the semantic vehicle partial segment results. An overview of our framework can be found in Fig.~\ref{fig:framework}.

To bridge the data gap of vehicle segmentation, in this work, we propose a new large-scale benchmark dataset, termed VehicleSeg10K. It contains 11,665 images, 13 vehicle part categories, collected from diverse scenarios and fully reflects the key challenges, including different views, illuminations, weathers, categories, etc. These images are carefully annotated and refined to support the validation of our SAV framework. These images are divided into training and testing subsets, which contain 8,596 and 3,069 images, respectively. In addition, we train and evaluate 18 baseline segmentation models to build a comprehensive benchmark. Both the dataset and pre-trained baselines will be open-sourced.

To sum up, the contributions of this paper can be summarized as the following three aspects: 

$\bullet$ We introduce a novel vehicle part segmentation framework based on pre-trained foundation model, termed SAV (Segment Anything Vehicle), which integrates a vehicle partial knowledge graph and vision context sample augmentation strategy. It leverages a vehicle part ontology to effectively capture spatial and geometric relationships among vehicle components, while also utilizing retrieved visual context to refine segmentation performance.

$\bullet$ We propose a large-scale benchmark dataset for vehicle segmentation, termed VehicleSeg10K, which contains 11665 images collected under different views, categories, illuminations, weathers, etc. Eighteen state-of-the-art algorithms are evaluated on this dataset, which builds a solid foundation for vehicle part segmentation.

$\bullet$ Extensive experiments on multiple vehicle segmentation datasets demonstrating that our approach substantially outperforms existing methods in both segmentation accuracy and part-level semantic consistency.

\section{Related Work}

In this section, we review research related to SAM-based Segmentation, Knowledge Graph, and Context Sample Augmentation. More related works can be found in the following surveys~\cite{yuan2024visionkg} and paper list~\footnote{\url{https://github.com/Event-AHU/Knowledge_Enhanced_Learning}}.

\subsection{SAM-based Segmentation}
The Segment Anything Model (SAM)~\cite{kirillov2023segment}, trained on an unprecedented dataset of over a billion masks, demonstrates remarkable zero-shot transfer capabilities, revolutionizing the field of image segmentation. Despite its impressive performance, SAM's fundamental limitation lies in its lack of semantic understanding, requiring explicit prompts (points, boxes, or rough masks) to generate meaningful segmentation results, thereby restricting its application in fully automated scenarios. Several approaches have been proposed to address these limitations. Semantic-SAM~\cite{li2307semantic} enables text-driven segmentation by integrating CLIP embeddings into the SAM architecture. With a focus on input flexibility, SEEM~\cite{zou2023segment} extends SAM with multimodal prompting capabilities. Similarly, PerSAM~\cite{zhang2023personalize} improves model adaptability through reference examples, though it still relies on example masks for initialization. HQ-SAM~\cite{ke2023segment} improves segmentation quality through supplementary processing modules while maintaining the requirement for explicit prompts.

To address computational efficiency concerns, FastSAM~\cite{zhao2023fast} replaces SAM's transformer architecture with a CNN-based approach, achieving comparable performance while being significantly faster. EfficientSAM~\cite{xiong2024efficientsam} introduces SAM-leveraged masked image pretraining to create lightweight models with reduced parameters. MobileSAM focuses on mobile deployment optimization, while recent SAM 2~\cite{ravi2024sam} extends capabilities to video segmentation. In the temporal domain, SAT~\cite{cheng2023segment} extends the application scope to video segmentation but still requires initial frame prompts. Although these extensions enhance SAM's capabilities, they remain dependent on user-provided explicit prompts, limiting their potential for automatic deployment. Our work addresses this fundamental constraint through a dual-stream prototype learning framework, where class prototypes derived from vehicle structure and context serve as implicit prompts.

\subsection{Knowledge Graph}
Knowledge graphs encode entities and their relationships, providing structured semantic representations that enhance visual understanding systems. In computer vision, integrating such relational context with visual features improves performance in fine-grained recognition and segmentation tasks, where semantic relationships complement ambiguous or limited visual information. Marino et al.~\cite{marino2016more} demonstrated the effectiveness of integrating knowledge graphs into visual recognition systems using graph search neural networks. Building upon this foundation, Chen et al.~\cite{chen2018knowledge} introduced knowledge embedding representation learning, jointly optimizing visual representation learning and knowledge graph embeddings. Recent advancements further improve integration capabilities; Wang et al.~\cite{wang2019kgat} proposed graph attention networks for knowledge-aware feature enhancement, while Liu et al.~\cite{liu2020attribute} developed knowledge propagation networks that transmit information along graph edges. These methods have shown promising results in classification tasks but remain underexplored for segmentation applications and have not been adequately adapted to capture the unique structural relationships present in vehicles. 
Our method bridges this gap by adopting a graph attention architecture to enrich textual embeddings of vehicle components with structural relationships, capturing topological information that enhances segmentation accuracy.

\subsection{Context Sample Augmentation}
Context-aware learning represents a key advancement in computer vision, enabling models to adapt to specific image conditions rather than relying solely on global representations. For segmentation tasks, incorporating contextual information improves boundary accuracy and enhances differentiation between visually similar categories, which is a critical aspect for vehicle component segmentation. 
The concept of prototype-guided segmentation was pioneered by Dong and Xing~\cite{dong2018few} in few-shot learning, where class prototypes constructed from support examples classify query image regions. 
Wang et al.~\cite{wang2019panet} extended this approach through prototype alignment networks, performing bidirectional feature alignment between support and query images. 
Li et al.~\cite{li2021adaptive} further advanced this direction through adaptive prototype learning, dynamically adjusting prototypes based on query image features. 
To integrate global context, Tian et al.~\cite{tian2020prior} proposed prior-guided feature enrichment networks. 
Within transformer-based architectures, Wang et al.~\cite{wang2021transformer} and Xie et al.~\cite{xie2021segmenting} demonstrated the effectiveness of attention mechanisms in capturing contextual dependencies.
While these methods are effective for general object segmentation, they typically require manually annotated examples for each new category and rarely integrate contextual information with structural knowledge. Our method enhances context learning by dynamically refining class representations based on visual content, synergistically working with our semantic graph to achieve autonomous segmentation of diverse vehicle appearances.

\section{Methodology}\label{sec:method}

\begin{figure*}[t]
\centering
\includegraphics[width=0.95\textwidth]{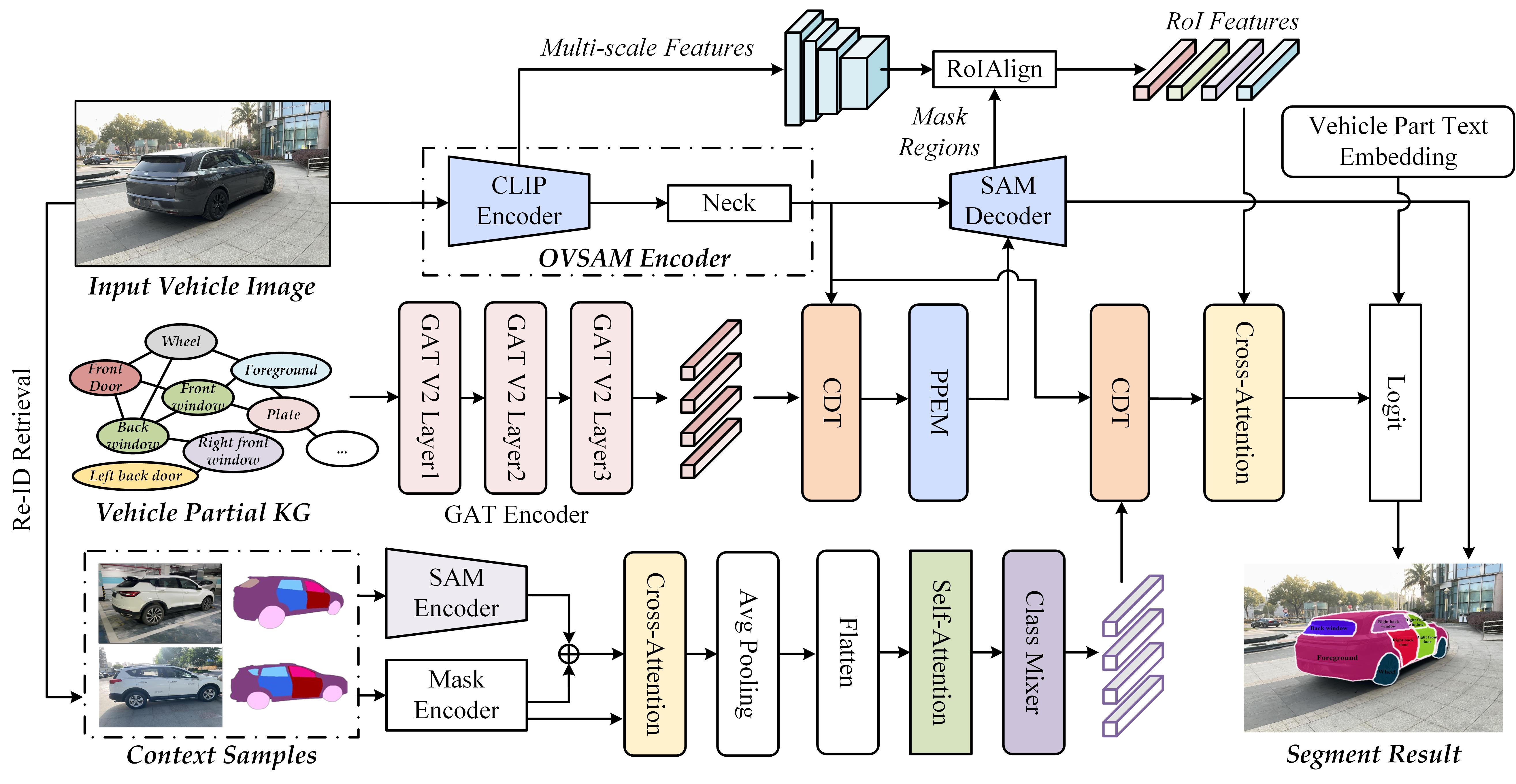}
\caption{Framework of our Prompt-Free Vehicle Part Segmentation approach. Our method transforms SAM into an autonomous vehicle part segmentation system through a dual-stream prototype learning mechanism. The upper stream processes the target vehicle through a modified SAM decoder to support multi-class output. The middle stream employs a knowledge graph to enhance part embeddings with structural relationships, while the lower stream utilizes a Re-ID model to extract visual part prototypes from similar reference vehicles. By combining these structure-aware textual and visual prototypes, our approach eliminates the need for explicit prompts while simultaneously segmenting all vehicle parts with proper identification.}
\label{fig:framework}
\end{figure*}

\subsection{Overview}
As shown in Figure~\ref{fig:framework}, we propose a vision-semantic context driven universal vehicle segmentation framework, termed SAV, which aims to effectively capture the spatial-geometric relationships between vehicle parts by constructing a vehicle part ontology and enhance existing segmentation performance through visual context retrieval, (i.e., acquiring vehicle images from similar viewpoints). Specifically, for an input vehicle image, we first employ the OVSAM~\cite{yuan2024ovsam} encoder to extract multi-scale feature representations, where the encoder integrates a frozen CLIP~\cite{radford2021learning} visual encoder and a multi-scale feature fusion neck network.
Simultaneously, to improve SAM's understanding of spatial relationships between vehicle parts, we construct a vehicle part knowledge graph using part label features extracted by the CLIP text encoder as entities and physical adjacency relationships between parts as edges. We then adopt the GATv2~\cite{brody2021attentive} graph attention network to encode the vehicle graph and fuse it with visual features through CDT and PEEM operations. The original visual features and VPKG-enhanced representations are fed into the SAM decoder to output region-of-interest features.
Furthermore, to address the challenge of distinguishing visually similar parts across different vehicle models and viewpoints, we retrieve vehicle context samples using the TransReid~\cite{he2021transreid} vehicle re-identification model and inject them into the decoding stage to further enrich the representations. Finally, semantic-aware vehicle part segmentation results are obtained through cross-attention and logit operations.

\subsection{Preliminary: SAM-Based Segmentation}

The Segment Anything Model (SAM) revolutionized interactive segmentation through its prompt-driven architecture, employing a Vision Transformer backbone to encode input images and extract visual features. The model processes user inputs (points, boxes, or masks) through a prompt encoder to generate prompt embeddings, which are subsequently fed to a mask decoder that produces binary segmentation masks. However, SAM exhibits several inherent limitations when applied to vehicle part segmentation tasks.

First, SAM requires explicit user prompts, making it impractical for autonomous scenarios where real-time processing must occur without human supervision. Second, the model produces only binary masks rather than multi-class outputs necessary for comprehensive part segmentation. Third, SAM lacks semantic understanding of the content it segments, struggling to distinguish between different vehicle parts that share similar visual characteristics but serve distinct functions.

To address these fundamental constraints, we modify SAM's operational paradigm by eliminating the dependency on user prompts and expanding the mask decoder to support multi-class segmentation. Our modified architecture employs an OVSAM encoder to extract multi-scale feature representations from input images $\mathbf{I} \in \mathbb{R}^{640 \times 480 \times 3}$, which are subsequently processed through a transformer-based neck network to produce enhanced feature maps $\mathbf{F} \in \mathbb{R}^{256 \times 64 \times 64}$. Instead of relying on user-provided prompts, we generate implicit prompts through our dual prototype mechanism, creating prototype-guided embeddings $\mathbf{P} \in \mathbb{R}^{13 \times 256}$ that replace traditional user inputs.

The mask decoder architecture is fundamentally redesigned to accommodate simultaneous segmentation of all vehicle parts. We expand the number of mask tokens from three to the total number of vehicle part classes (13), with each class having a dedicated hypernetwork MLP for mask prediction. This architectural modification enables the generation of class-specific masks $\mathbf{M} = \{M_1, M_2, \ldots, M_{13}\}$ where each $M_i \in \mathbb{R}^{1024 \times 1024}$, allowing simultaneous segmentation of all potential parts in a single forward pass:

\begin{equation}
\mathbf{M} = \text{Decoder}(\mathbf{F}, \mathbf{P}),
\end{equation}
where $\mathbf{M} \in \mathbb{R}^{n \times 1024 \times 1024}$ represents the set of class-specific masks for all vehicle parts, $\mathbf{P} \in \mathbb{R}^{n \times 256}$ contains the prototype-guided prompts generated by our dual prototype mechanism, and $n$ is the total number of vehicle part classes in our framework.

\subsection{Vehicle Partial Knowledge Graph}
Unlike articulated objects with flexible joints, vehicle components have fixed spatial relationships, and these rigid spatial constraints can provide rich structural priors for part segmentation. To better leverage this spatial relationship, we constructed a vehicle partial knowledge graph $G = (V, E, W)$, where $V={v_1, v_2, ..., v_n}$ represents the set of vehicle part nodes corresponding to our 13 semantic categories, $E$ denotes the edges connecting spatially adjacent parts, and $W \in \mathbb{R}^{n \times n}$ represents the adjacency matrix with learned edge weights. 

\begin{figure}
    \centering
    \includegraphics[width=0.5\textwidth]{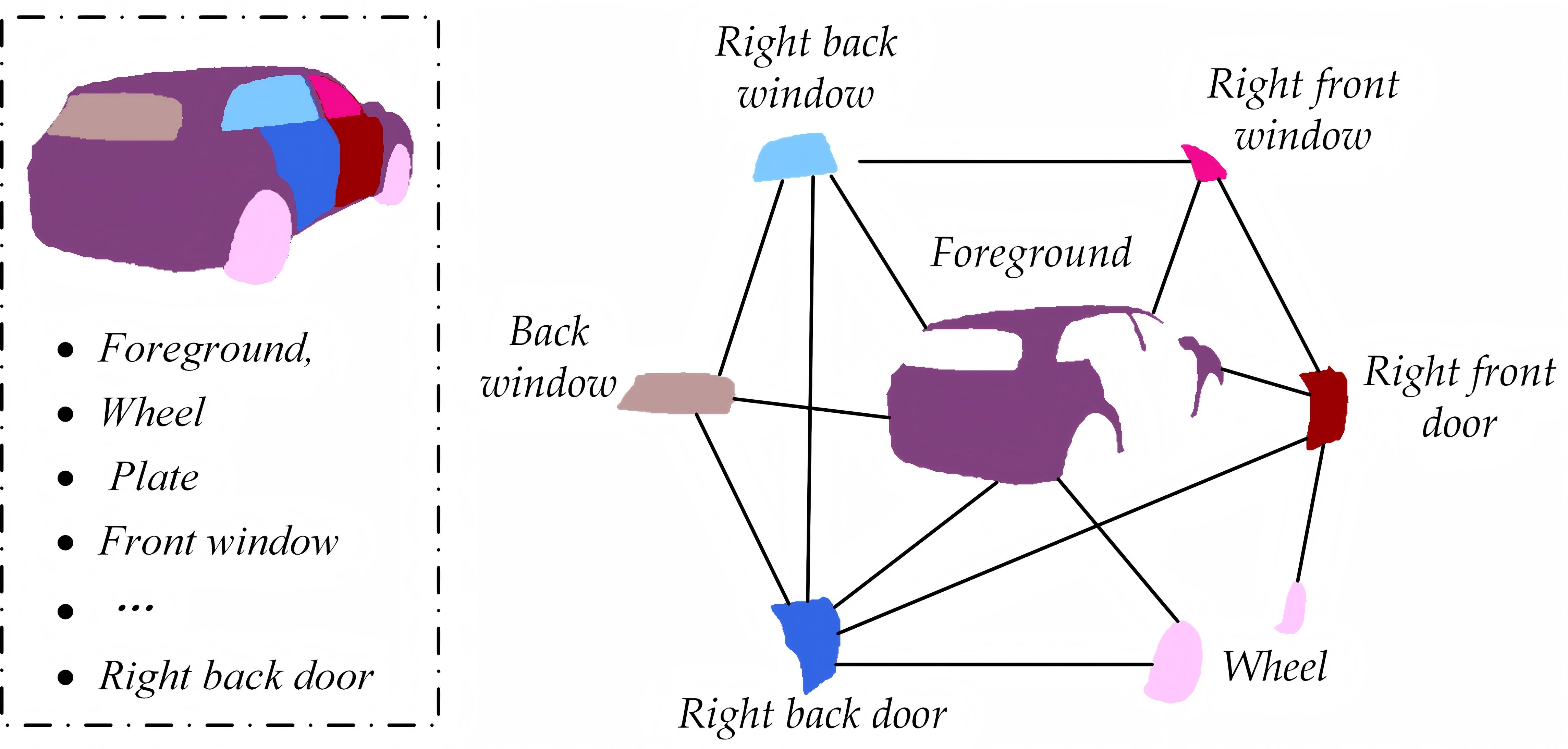}
    \caption{An overview of the vehicle partial knowledge graph.} 
    \label{fig:KGgraph}
\end{figure}

Specifically, we employ the CLIP text encoder to extract text embeddings of category words as nodes. The edge connections are established based on the physical structure of the vehicle, such as the left front door connects to the left front window, left back door, and foreground (vehicle body), while maintaining no connection to right-side components due to the vehicle's inherent bilateral symmetry. This approach ensures that the graph structure accurately represents vehicle topology without requiring complex geometric relationship modeling. The complete graph structure is shown in Figure~\ref{fig:KGgraph}. For each pair of connected parts $(i,j) \in E$, we compute the edge weight $W_{ij}$ based on their co-occurrence frequency in the training dataset:
\begin{equation}
W_{ij} = \frac{\text{Count}(i \cap j)}{N},
\end{equation}
where $\text{Count}(i \cap j)$ represents the number of training samples where both parts $i$ and $j$ are present simultaneously, and $N$ is the total number of training samples. This statistical approach captures the likelihood of parts appearing together, providing valuable contextual information for the segmentation process.

Initially, each vehicle part class $i$ is represented by a textual embedding $\mathbf{T}_i \in \mathbb{R}^{1\times768}$ generated using CLIP's text encoder. We then construct a graph \( G = (V, E, W) \), where \( V \) is the set of vertices representing the $n$ vehicle part classes, \( E \) is the set of edges connecting physically adjacent parts, and \( W \in \mathbb{R}^{n \times n} \) represents edge weights based on co-occurrence statistics derived from the training dataset. 

We enhance these initial textual embeddings $\mathbf{T}$ through a four-layer Graph Attention Network (GATv2) that enables attention-based information flow between adjacent parts. The four-layer architecture is empirically determined through systematic experimental analysis detailed in Section V-D.:
\begin{equation}
\mathbf{P}_t = \text{GATv2}(\mathbf{T}, E, W),
\end{equation}
where $\text{GATv2}$ employs the following attention mechanism for each layer:
\begin{equation}
\alpha_{ij} = \frac{\exp\left(\mathbf{a}^T \cdot \text{LeakyReLU}(\mathbf{W}_l\mathbf{h}_i + \mathbf{W}_r\mathbf{h}_j)\right)}{\sum_{k \in \mathcal{N}_i} \exp\left(\mathbf{a}^T \cdot \text{LeakyReLU}(\mathbf{W}_l\mathbf{h}_i + \mathbf{W}_r\mathbf{h}_k)\right)},
\end{equation}
\begin{equation}
\mathbf{h}'_i = \sigma\left(\sum_{j \in \mathcal{N}_i} \alpha_{ij} \mathbf{W}_r \mathbf{h}_j\right),
\end{equation}
where $\alpha_{ij}$ represents the attention coefficient between nodes $i$ and $j$, $\mathbf{W}_l$ and $\mathbf{W}_r$ denote the learnable left and right weight matrices for linear transformation, $\mathbf{h}_i$ and $\mathbf{h}_j$ are the feature vectors of nodes $i$ and $j$ respectively, $\mathbf{a}$ is the learnable attention vector, $\mathcal{N}_i$ denotes the neighborhood of node $i$. This design choice combines the benefits of explicit structural knowledge through hard-coded adjacencies with data-driven insights through co-occurrence statistics, enabling the model to capture both the spatial constraints and statistical dependencies of vehicle parts. Our implementation utilizes a multi-head attention mechanism with 4 heads in the first two layers, followed by a single-head output layer, enabling the model to jointly attend to different feature subspaces.

The generated prototype \(\mathbf{P}_t \in \mathbb{R}^{n \times 256}\) integrates structural constraints while preserving part-specific semantic information. To integrate these prototypes with image features, we employ a Content-Dependent Transfer (CDT) module that implements a cross-attention mechanism with positional encoding:
\begin{equation}
\tilde{\mathbf{P}}_t = \text{CDT}_t(\mathbf{F}, \mathbf{P}_t) = \text{CrossAttn}(\mathbf{P}_t, \text{PE}(\mathbf{F}), \mathbf{F}),
\end{equation}
where \(\text{PE}\) denotes the positional encoding of spatial features. This adaptability enables the prototypes to focus on relevant image regions while maintaining their semantic integrity.

The adapted textual prototypes are processed through a Prototype-Prompt Encoder for Multi-Class segmentation (PPEM) that generates both dense and sparse embeddings:
\begin{equation}
\mathbf{D}, \mathbf{S} = \text{PPEM}(\mathbf{F}', \tilde{\mathbf{P}}_t),
\end{equation}
where $\mathbf{F}'$ represents flattened image features, $\mathbf{D}$ indicates dense embeddings, and $\mathbf{S}$ represents sparse embeddings.

The PPEM first computes similarity between image features and prototypes, then generates activated features through similarity-weighted enhancement. For each image feature vector $\mathbf{F}'_i$ and each graph-enhanced prototype $\tilde{\mathbf{P}}_{t,c}$ (where $c$ indexes the class), the activation is computed as:
\begin{equation}
\mathbf{A}_{i,c} = \mathbf{F}'_i + \mathbf{F}'_i \odot (\mathbf{F}'_i \cdot \tilde{\mathbf{P}}_{t,c}).
\end{equation}
These activated features are transformed into dense and sparse embeddings through parallel pathways to guide the mask generation process.In the dense pathway, class-specific features undergo MLP processing before being concatenated and convolved into a unified spatial representation. Concurrently, the sparse pathway transforms activated features into token embeddings using an MLP, then enhances them with class-specific positive or negative embeddings based on class presence.

\subsection{Context Sample Augmentation}

While graph-enhanced textual prototypes effectively capture semantic information and structural relationships between vehicle parts, they face limitations in handling appearance variations across different vehicle types, viewpoints, and lighting conditions. Vehicle parts can exhibit dramatically different visual characteristics depending on the vehicle model and viewing conditions, creating ambiguity in semantic prediction. To address this limitation in semantic mask identification, we introduce a context-aware visual prototype module that leverages similar vehicle examples as references, providing appearance-specific guidance to enhance the semantic classification accuracy of the generated masks.

\begin{figure}
\centering
\includegraphics[width=1\linewidth]{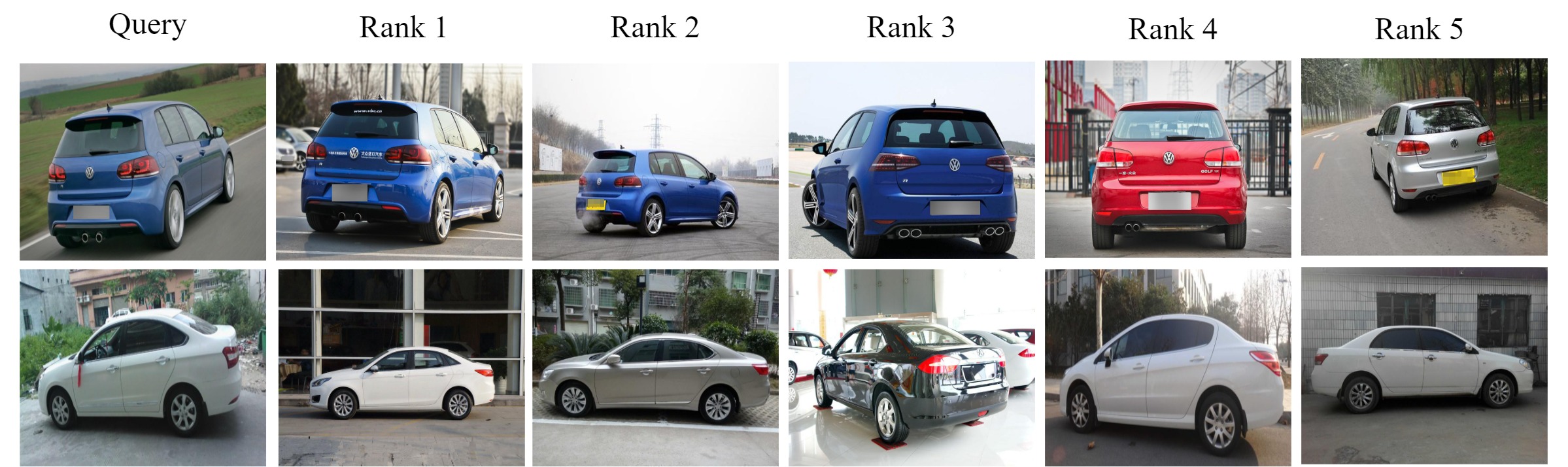}
\caption{Visualization of the Results of Vehicle Re-Identification (ReID)}
\label{fig:reid}
\end{figure}

For each input image $I$, we employ a vehicle re-identification (ReID) model to retrieve visually similar vehicles from a reference database. Figure~\ref{fig:reid} illustrates the effectiveness of our ReID-based retrieval mechanism in identifying visually similar vehicle examples. Our ReID model is built upon a robust foundation established through VehicleMAE pre-training on one million vehicle images, with the visual backbone of TransReID initialized using these pre-trained weights and subsequently fine-tuned on vehicle re-identification datasets. During the retrieval phase, when processing a query vehicle, the ReID model extracts features $\mathbf{f} \in \mathbb{R}^{768}$ and computes similarity with gallery vehicles using cosine similarity in the L2-normalized feature space:

\begin{equation}
\text{sim}(I_q, I_g) = \frac{f_q \cdot f_g}{\|f_q\| \|f_g\|},
\end{equation}

\begin{equation}
\{I'_1, I'_2, \ldots, I'_k\}, \{M'_1, M'_2, \ldots, M'_k\} = \text{ReID}(I, \mathcal{G}),
\end{equation}
where $I$ represents the input image, $\mathcal{G}$ represents the gallery database, $I'_i \in \mathbb{R}^{3 \times H \times W}$ are the retrieved similar vehicle images, and $M'_i \in \mathbb{R}^{13 \times H \times W}$ are their corresponding segmentation masks. The retrieval process typically returns $k=2$ reference examples to balance information richness with computational efficiency.

The retrieved reference examples serve as visual prompts for generating class-specific prototypes through a specialized PromptImageEncoder. This encoder processes multi-modal inputs including points, boxes, and masks from reference examples, transforming them into discriminative embeddings. The encoder handles both sparse signals (points, boxes) and dense signals (masks) into a unified embedding space, with masks being processed through a hierarchical convolutional pathway that progressively reduces spatial dimensions from the original mask resolution to a $64 \times 64$ feature representation while increasing feature depth to 256 dimensions.

The prototype generation involves a hierarchical multi-head attention mechanism operating on three distinct dimensions. The attention operations process embeddings $\mathbf{E}_s \in \mathbb{R}^{2 \times 13 \times N \times 256}$ through class attention $A_c$, example attention $A_e$, and class-example attention $A_{ce}$:

\begin{equation}
\hat{E} = \Phi_{\text{proj}}(A_{\text{ce}}(A_{\text{e}}(A_{\text{c}}(\Phi_{\text{proj}}^{-1}(E_{s}))))).
\end{equation}
Each attention operation follows the standard formulation:
\begin{equation}
\text{Attention}(Q, K, V, M_{\text{mask}}) = \text{softmax}\left(\frac{QK^T}{\sqrt{d}} + M_{\text{mask}}\right)V,
\end{equation}
where $M_{\text{mask}}$ is derived from the support flags $F$ to mask invalid classes. After attention processing, the resulting embeddings are pooled and normalized to form class-specific prototypes $\mathbf{P}_v \in \mathbb{R}^{B \times 13 \times 768}$:

\begin{equation}
\mathbf{P}_v = \frac{\sum_{m=1}^{M} \hat{E}_m \odot F_m}{\sum_{m=1}^{M} F_m + \epsilon}.
\end{equation}

Unlike textual prototypes that directly guide the mask decoder, visual prototypes enhance classification through post-segmentation refinement. Both prototype streams initially undergo parallel Content-Dependent Transfer (CDT) processing, where cross-attention mechanisms adapt prototype features to the specific image content. The visual prototypes $\tilde{\mathbf{P}}_v$ are transformed through this process to align with the current image features. After initial mask prediction, we extract ROI features from predicted regions using a Feature Pyramid Network (FPN) that processes backbone features at four scales [384, 768, 1536, 3072] into unified 256-dimensional representations. These ROI features, obtained through alignment with mask-derived bounding boxes, are then enhanced via cross-attention with visual prototypes. In this attention mechanism, ROI features serve as queries while visual prototypes function as keys and values, establishing weighted relationships that identify relevant prototype dimensions for each part. The resulting attention-modulated features enable accurate part classification by prioritizing visual characteristics that match detected regions while suppressing irrelevant information, ultimately facilitating precise distinction between visually similar parts based on their spatial context and appearance characteristics.

\begin{figure*}[t]
    \centering
    \includegraphics[width=1\linewidth]{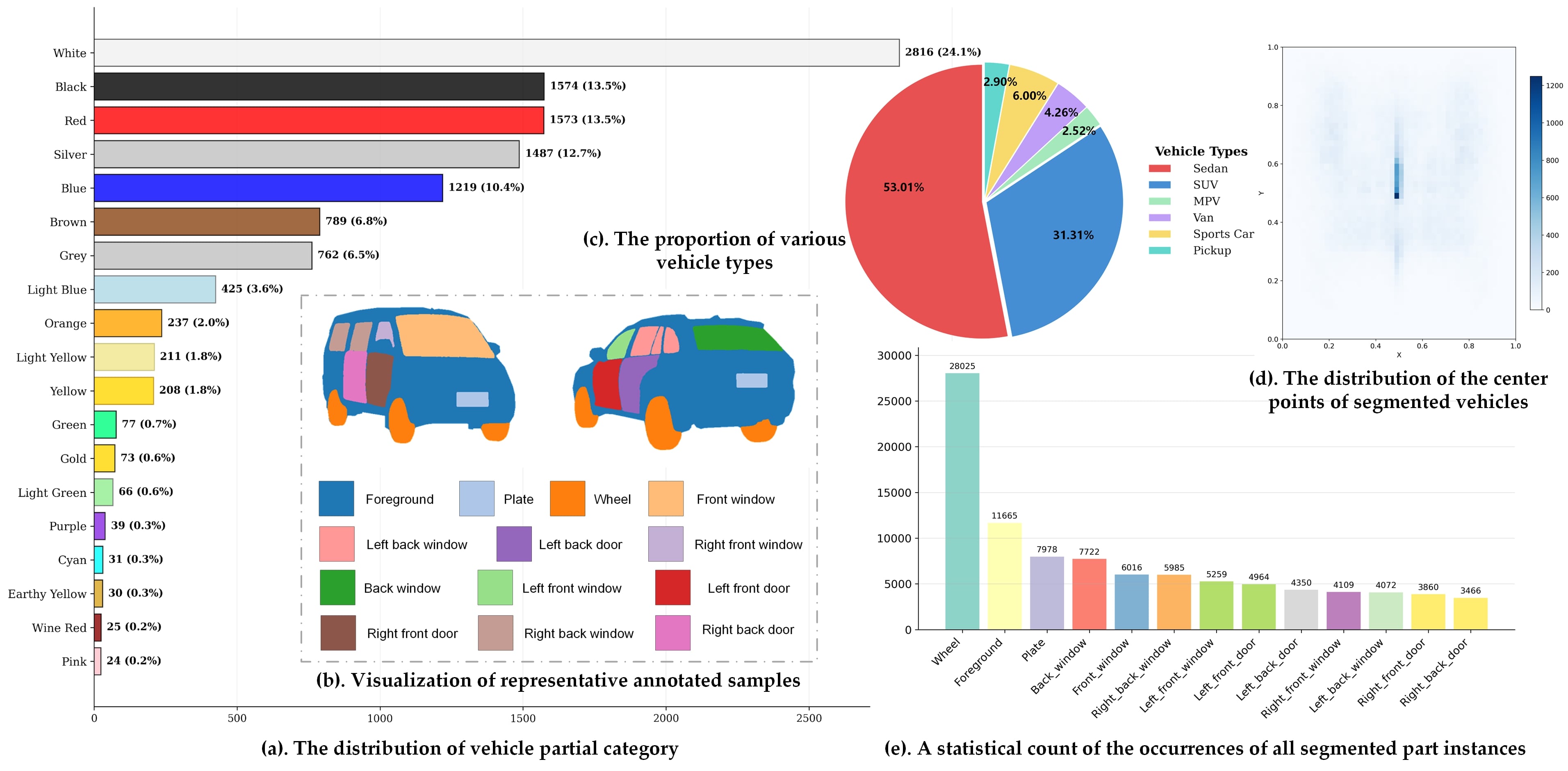}
    \caption{\textbf{Visualization of All Annotation Information in the Dataset.} (a) Distribution of vehicle colors across the entire dataset; (b) Visualization examples of representative annotated vehicle types in the dataset; (c) Proportion of different vehicle types in the entire dataset; (d) Heatmap showing the distribution of vehicle center points; (e) Statistical count of annotated instances for different vehicle parts.}
    \label{fig:Annotation}
\end{figure*}

\subsection{Loss Function}
Our model training employs three specialized loss components that work together to ensure accurate segmentation and classification.
The mask loss employs binary cross-entropy to measure pixel-wise accuracy between predicted masks and ground truth:
\begin{equation}
\mathcal{L}_{\text{mask}} = -\frac{1}{N}\sum_{i=1}^{N}[y_i\log(p_i) + (1-y_i)\log(1-p_i)],
\end{equation}
where $y_i$ represents the ground truth mask value (0 or 1) at pixel $i$, $p_i$ denotes the predicted probability, and $N$ is the total number of sampled points. This loss encourages accurate pixel-level predictions, particularly at part boundaries where precision is critical.
The Dice loss addresses potential class imbalance and focuses on segmentation quality by measuring spatial overlap between predicted and ground truth masks:
\begin{equation}
\mathcal{L}_{\text{dice}} = 1 - \frac{2\sum_{i}^{N}p_i y_i}{\sum_{i}^{N}p_i^2 + \sum_{i}^{N}y_i^2 + \epsilon},
\end{equation}
where $\epsilon$ is a small constant for numerical stability. This loss is particularly effective for smaller vehicle parts that occupy fewer pixels but are equally important for comprehensive segmentation.
The classification loss ensures accurate identification of each vehicle part through weighted cross-entropy:
\begin{equation}
\mathcal{L}_{\text{cls}} = -\sum_{c=1}^{C} w_c\sum_{j=1}^{M} y_{j,c}\log(p_{j,c}),
\end{equation}

These losses are combined into a unified training objective:
\begin{equation}
\mathcal{L} = \lambda_1\mathcal{L}_{\text{mask}} + \lambda_2\mathcal{L}_{\text{dice}} + \lambda_3\mathcal{L}_{\text{cls}}
\end{equation}
with weights $\lambda_1=5.0$, $\lambda_2=5.0$, and $\lambda_3=2.0$ determined empirically to balance boundary precision, region coverage, and classification accuracy. During training, we employ uncertainty-guided point sampling that focuses on regions with high prediction uncertainty, improving segmentation precision for adjacent vehicle parts with similar appearances.

\section{Benchmark Dataset} \label{sec:experiments}

\begin{table*}
\centering
\caption{Comparison of existing car part datasets.}
\label{tab:car_part_datasets}
\begin{tabular}{l|c|c|c|c|c|c|c|c c|c}
\hline
\multirow{2}{*}{\textbf{Dataset}} & \multirow{2}{*}{\textbf{Year}} & \multirow{2}{*}{\textbf{Part IDs}} & \multirow{2}{*}{\textbf{Vehicle Types}} & \multirow{2}{*}{\textbf{Images}} & \multirow{2}{*}{\textbf{Resolution}} & \multirow{2}{*}{\textbf{Viewpoints}} & \multirow{2}{*}{\textbf{Real}} & \multicolumn{2}{c|}{\textbf{Scenarios}} & \multirow{2}{*}{\textbf{Weather Adverse}} \\
& & & & & & & & \textbf{Day} & \textbf{Night} & \\
\hline
UDA-Part~\cite{liu2021cgpart} & 2021 & 31 & 4 & 3,862 & 369-800$^2$ & Fixed &  &  &  &  \\
3DRealCar-Part~\cite{du20243drealcar} & 2024 & 13 & 6 & 10,219 & 1920$\times$1440 & Horizontal only & \checkmark & \checkmark &  &  \\
VehicleSeg10K (Ours) & 2025 & 13 & 6 & 11,665 & 229-1411$^2$ & Multi-angle & \checkmark & \checkmark & \checkmark & \checkmark \\
\hline 
\end{tabular}
\end{table*}

\subsection{Protocols}
We aim to provide a comprehensive platform for training and evaluation of vehicle part segmentation models. When constructing our VehicleSeg10K benchmark dataset, we follow the following protocols: 

\textbf{1) Large-scale}: With the rapid advancement of autonomous driving technology and the increasing demand for fine-grained vehicle understanding, large-scale datasets demonstrate crucial importance for robust model development. Our dataset contains 11,665 vehicle images with 97,471 part instances across 13 categories, providing substantial training data that significantly exceeds existing benchmarks in both scale and annotation density.

\textbf{2) Diversity}: During the collection process, we anticipated potential challenges in real-world deployment scenarios and systematically incorporated diverse factors that significantly impact vehicle perception performance. As shown in Table~\ref{tab:car_part_datasets}, our dataset addresses critical limitations in existing datasets where UDA-Part's synthetic nature lacks environmental variations and employs fixed viewpoints, while 3DRealCar-Part lacks comprehensive viewing angle coverage and adverse weather scenarios. More in detail, adverse weather conditions including rain, fog, snow, and dust storms, challenging outdoor nighttime illumination scenarios, multi-angle viewpoints encompassing elevated and ground-level perspectives with comprehensive vertical angle variations, multi-resolution conditions ranging from $229 \times 202$ to $1411 \times 1302$ pixels, and diverse vehicle types including Sedan, SUV, MPV, Van, Sports Car, and Pickup are all systematically considered when collecting these vehicle images. These challenging factors address critical gaps in existing datasets, providing the first large-scale vehicle part segmentation dataset that simultaneously addresses scale, diversity, and real-world deployment challenges.

\textbf{3) Fine-grained Parts}: We focus on enhancing the segmentation capability for fine-grained vehicle components, which are essential for advanced automotive applications but often challenging due to similar visual appearances and complex spatial relationships. The dataset encompasses 13 distinct component categories, namely Foreground, Wheel, Plate, Front window, Back window, Left front window, Left front door, Left back window, Left back door, Right front window, Right front door, Right back window, and Right back door. This comprehensive part coverage provides semantically meaningful annotations that enable detailed vehicle understanding required for damage assessment, automated parking assistance, and intelligent transportation systems.

\begin{figure*}
\centering
\includegraphics[width=1\linewidth]{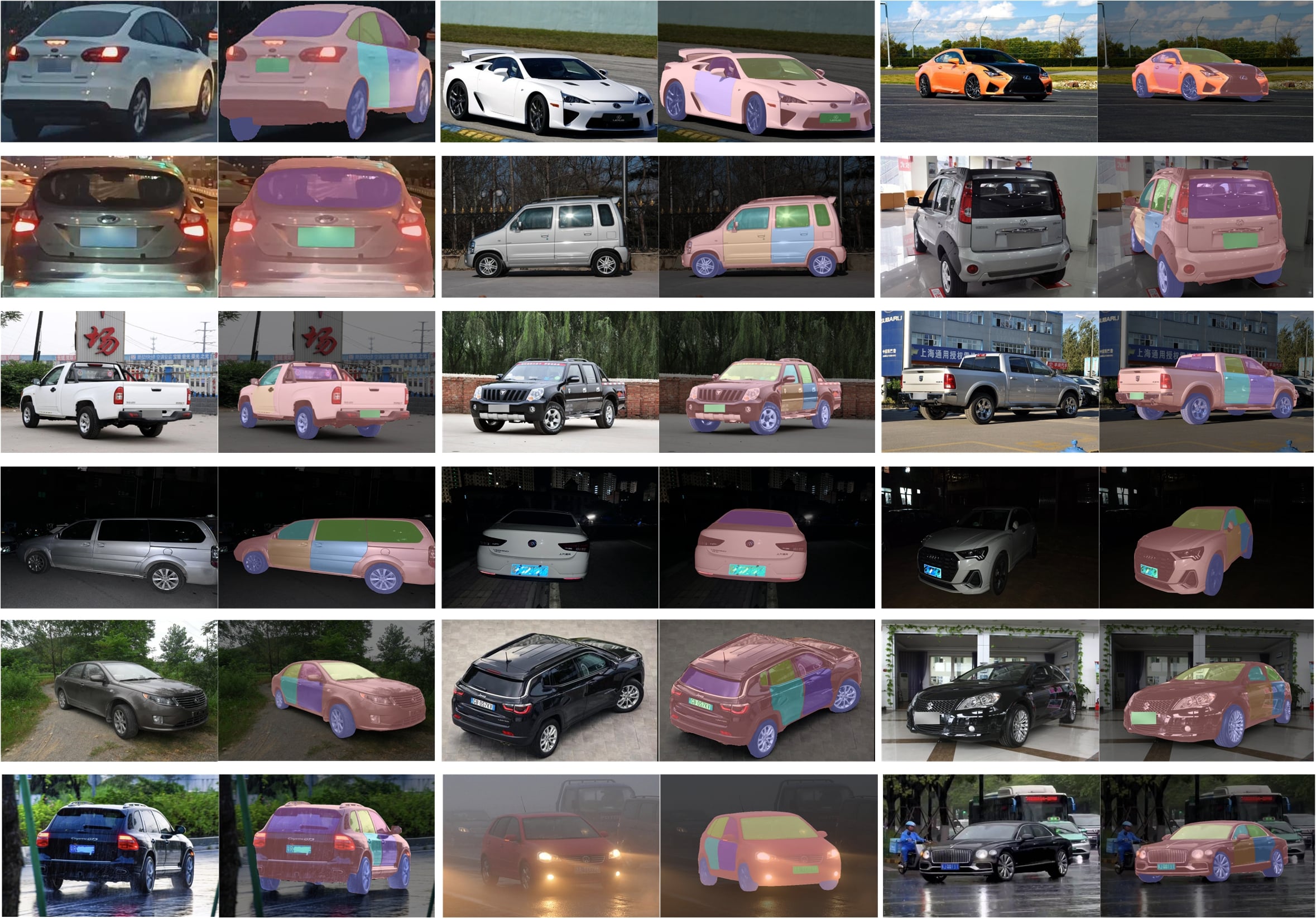}
\caption{Representative samples of our proposed VehicleSeg10K dataset.}
\label{fig:showAnnotation}
\end{figure*}

\subsection{Data Collection and Annotation}

Vehicle images are collected through web crawling with data privacy protection measures. Image resolutions range from $229 \times 202$ to $1411 \times 1302$ pixels, reflecting diverse capture conditions. We employ pixel-level precise segmentation masks with unique identifiers for each part category.

The annotation process involves student annotators under expert supervision. Quality control is maintained through a multi-stage validation mechanism where specialized reviewers assess annotation quality, followed by final expert validation. Annotations are stored in standard formats compatible with deep learning frameworks.Figure~\ref{fig:showAnnotation} presents representative samples from our VehicleSeg10K dataset, demonstrating the diversity of vehicle types, environmental conditions (including nighttime and adverse weather), and multi-angle viewpoints. The color-coded segmentation masks precisely delineate the 13 vehicle part categories across various challenging real-world scenarios.

\subsection{Statical Analysis}
Our constructed dataset comprises 11,665 vehicle images with 97,471 part annotations across 13 categories. As shown in Figure~\ref{fig:Annotation}, this figure comprehensively presents multiple statistical information, including part distribution, vehicle type distribution, color distribution, and vehicle center point distribution heatmap. Wheels, as the most common component, occupy a prominent position in the figure, followed by foreground and various window/door categories. The figure also reveals the distribution of six major vehicle types, with sedans and SUVs constituting the majority. The color distribution section demonstrates the authentic distribution characteristics of vehicles in the real world. The vehicle center point distribution heatmap statistically analyzes the distribution of overall center points of vehicles to be segmented across the entire dataset, providing intuitive evidence for studying positional preferences of vehicles within images. The entire dataset is divided into 8,596 training images and 2,150 testing images to support model training and evaluation processes.

\subsection{Benchmark Baselines} 
Based on our proposed VehicleSeg10K dataset, we retrain and evaluate 18 baseline models to build a comprehensive benchmark. More in detail, 
DeepLabV3~\cite{chen2017rethinking}, DeepLabV3+~\cite{chen2018encoder}, DDRNet~\cite{hong2021deep}, PoolFormer~\cite{yu2022metaformer}, ConvNeXT~\cite{liu2022convnet}, SegNeXT~\cite{guo2022segnext}, MaskDINO~\cite{li2023mask}, 
GSS-gss~\cite{chen2023generative}, PEM~\cite{cavagnero2024pem}, FADC~\cite{chen2024frequency}, HieraHyp~\cite{weber2024flattening}, CGRSeg~\cite{ni2024context}, Spike2Former~\cite{lei2025spike2former},
gcnet~\cite{yang2025golden}, 
ProPETL~\cite{zhou2025propetl},
SegMAN~\cite{SegMAN},
CWSAM~\cite{pu2025classwise}, 
WPS-SAM~\cite{wu2024wps}.

\section{Experiment}\label{sec:experiments}

\subsection{Dataset and Evaluation Metric}

In the experiments, we evaluate our method on three vehicle segmentation datasets, including \textbf{3DRealCar-Part}, \textbf{UDA-Part}, and our newly proposed \textbf{VehicleSeg10K} dataset. A brief introduction to the first two datasets is given below. 

\noindent $\bullet$ \textbf{3DRealCar-Part} is a subset derived from the 3DRealCar dataset~\cite{du20243drealcar}, specifically designed for 2D semantic segmentation tasks. It comprises 7,965 training images and 2,254 validation images. This dataset includes 13 vehicle component categories, such as license plates, wheels, front windows, rear windows, left front windows, left front doors, left rear windows, left rear doors, right front windows, right front doors, right rear windows, right rear doors, as well as a foreground category representing the car body. High-quality segmentation masks are provided for all vehicle components, enabling precise evaluation of segmentation performance.  

\noindent $\bullet$ \textbf{UDA-Part}~\cite{liu2021cgpart} dataset originates from the vehicle category of the CGPart dataset, containing 3,090 training images and 772 testing images with annotations covering 31 distinct vehicle components. Both the training and testing images are synthetic, which presents a challenging domain adaptation scenario. This allows for the assessment of our model's ability to generalize from synthetic data to real-world scenarios.

We employ two standard metrics to evaluate our approach: 
\textbf{Mean Intersection over Union (mIoU)}: This metric quantifies segmentation precision by computing the average overlap between predicted and ground truth masks across all classes:
\begin{equation}
\text{mIoU} = \frac{1}{C}\sum_{c=1}^{C} \frac{TP_c}{TP_c + FP_c + FN_c}
\end{equation}
\textbf{Mean Accuracy (mAcc)}: This metric evaluates classification performance by calculating the average per-class pixel accuracy:
\begin{equation}
\text{mAcc} = \frac{1}{C}\sum_{c=1}^{C} \frac{TP_c}{TP_c + FN_c}
\end{equation}
These complementary metrics provide a comprehensive assessment: mIoU emphasizes boundary precision and region overlap, while mAcc focuses on class-specific identification accuracy.

\subsection{Implementation Details}

We train our model using the AdamW optimizer with a base learning rate of 0.0001, weight decay of 0.05, and batch size of 4. The learning rate schedule includes a linear warm-up phase for the first 500 training iterations, followed by step decay at training epochs 30, 40, and 45, with a decay factor (gamma) of 0.1. We employ different learning rate multipliers for different components: 0.1$\times$ for the backbone and 1.0$\times$ for embedding-related parameters.

For data augmentation, we resize images to 640$\times$480 while maintaining the aspect ratio and filter instances with mask areas smaller than 32 pixels. We train the model for 50 epochs with validation performed every 5 epochs. The loss weights are set to 2.0 for classification loss, 5.0 for mask loss, and 5.0 for Dice loss. During inference, we process input images at their native resolution to preserve fine details. The model simultaneously predicts confidence scores and mask logits for all potential vehicle parts. We apply a threshold of 0.5 to the sigmoid-activated mask logits to determine the presence and boundaries of each part class in the final segmentation output.

\begin{table*}[t]
\centering
\caption{Performance comparison with state-of-the-art methods on public vehicle segmentation benchmarks. Results measured by mean IoU (\%) and mean Accuracy (\%). Best results are highlighted in \textbf{bold}, and second-best results are \underline{underlined}.}
\label{tab:benchmark_comparison}
\footnotesize
\setlength{\tabcolsep}{3pt}
\begin{tabular}{c|l|l|l|cc|cc|cc|c}
\toprule
\multirow{2}{*}{\textbf{Index}} & \multirow{2}{*}{\textbf{Method}} & \multirow{2}{*}{\textbf{Backbone}} & \multirow{2}{*}{\textbf{Venue}} & \multicolumn{2}{c|}{\textbf{3DRealCar-Part}} & \multicolumn{2}{c|}{\textbf{UDA-Part}} & \multicolumn{2}{c|}{\textbf{VehicleSeg10K}} & \multirow{2}{*}{\textbf{Params}} \\
\cmidrule{5-10}
& & & & \textbf{mIoU} & \textbf{mAcc} & \textbf{mIoU} & \textbf{mAcc} & \textbf{mIoU} & \textbf{mAcc} & \\
\midrule
01 & DeepLabV3~\cite{chen2017rethinking} & ResNet-50 & CVPR'17 & 71.87 & 82.22 & 64.68 & 76.35 & 66.18 & 77.56 & 39.7 \\
02 & DeepLabV3+~\cite{chen2018encoder} & ResNet-50 & CVPR'18 & 72.04 & 83.53 & 65.23 & 77.28 & 66.23 & 77.66 & 54.4 \\
03 & DDRNet~\cite{hong2021deep} & DDRNet & T-ITS'22 & 64.23 & 77.05 & 63.66 & 76.75 & 64.12 & 75.90 & 28.2 \\
04 & PoolFormer~\cite{yu2022metaformer} & PF-M48 & CVPR'22 & 61.52 & 74.09 & 64.45 & 76.70 & 72.08 & 82.51 & 73.0 \\
05 & ConvNeXT~\cite{liu2022convnet} & ConvNeXt-XL & CVPR'22 & 61.13 & 74.01 & 63.69 & 76.39 & 72.58 & 82.50 & 350.0 \\
06 & SegNeXT~\cite{guo2022segnext} & MSCAN-L & NeurIPS'22 & 74.97 & \underline{84.70} & 65.42 & 77.58 & 64.26 & 76.26 & 48.9 \\
07 & MaskDINO~\cite{li2023mask} & Swin-L & CVPR'23 & 69.89 & 80.84 & \underline{69.13} & \underline{79.06} & 73.50 & 83.68 & 223.0 \\
08 & GSS-gss~\cite{chen2023generative} & Swin-L & CVPR'23 & 61.97 & 73.58 & 38.93 & 52.85 & 57.89 & 70.72 & 167.5 \\
09 & PEM~\cite{cavagnero2024pem} & ResNet-50 & CVPR'24 & 61.53 & 72.27 & 60.79 & 71.66 & 61.53 & 72.56 & 35.6 \\
10 & FADC~\cite{chen2024frequency} & ResNet-50 & CVPR'24 & 70.80 & 81.59 & 67.67 & 77.86 & 71.9 & 81.84 & 128.0 \\
11 & HieraHyp~\cite{weber2024flattening} & HRNetV2-w48 & CVPR'24 & 70.26 & 80.97 & 62.95 & 74.97 & 64.13 & 74.96 & 67.8 \\
12 & CGRSeg~\cite{ni2024context} & MSCAN-T & ECCV'24 & 66.85 & 79.72 & 65.04 & 75.82 & 70.01 & 80.67 & 35.7 \\
13 & Spike2Former~\cite{lei2025spike2former} & Meta-SF & AAAI'25 & 61.18 & 71.35 & 68.81 & 78.15 & 61.19 & 72.55 & 34.0 \\
14 & gcnet~\cite{yang2025golden} & GCBlocks & CVPR'25 & 68.48 & 83.03 & 66.39 & 78.59 & 73.40 & 83.11 & 34.2 \\
15 & ProPETL~\cite{zhou2025propetl} & Swin-L & ICLR'25 & 70.12 & 80.86 & - & - & 74.48 & \underline{84.14} & 316.2 \\
16 & SegMAN~\cite{SegMAN} & SegMAN & CVPR'25 & 62.23 & 73.62 & - & - & 62.69 & 73.97 & 51.8 \\
17 & CWSAM~\cite{pu2025classwise} & Sam Vit-b & J-STARS'25 & \underline{76.09} & 83.75 & - & - & \underline{76.90} & 82.38 & 104.96 \\
18 & WPS-SAM~\cite{wu2024wps} & Sam Vit-b & ECCV'24 & 75.71 & 82.57 & - & - & 74.27 & 83.04 & 101.6 \\
\midrule
19 & \textbf{SAV (Ours)} & ResNet-50 & - & \textbf{80.21} & \textbf{85.91} & \textbf{70.76} & \textbf{79.35} & \textbf{81.23} & \textbf{87.08} & 313.2 \\
\bottomrule
\end{tabular}
\end{table*}

\subsection{Comparison on Public Benchmark Dataset}
\noindent \textbf{Results on VehicleSeg10K Dataset.} Table~\ref{tab:benchmark_comparison} presents experimental results on our newly proposed VehicleSeg10K dataset. Among the methods listed, CWSAM and ProPETL achieve 76.90/82.38 and 74.48/84.14, respectively, while traditional methods like DeepLabV3+ achieve 66.23/77.66. Our model SAV outperforms all others with 81.23/87.08, demonstrating a significant improvement of +4.33 mIoU over the second-best method CWSAM. This substantial performance gain validates the effectiveness of our dual-prototype framework combining structural knowledge and visual context for vehicle part segmentation.

\begin{figure*}[t]
    \centering
    \includegraphics[width=\textwidth]{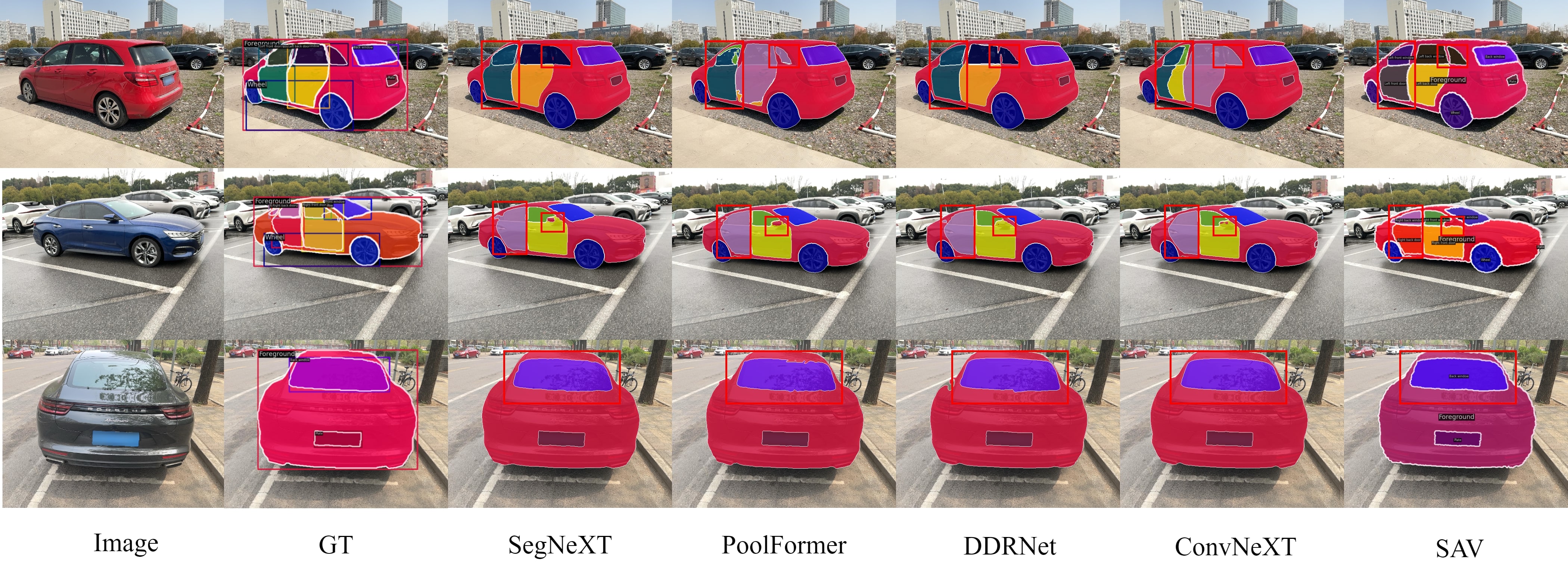}
    \caption{Performance Comparison of Semantic Segmentation Methods on 3DRealCar-Part Dataset}
    \label{fig:visual_results}
\end{figure*}

\noindent \textbf{Results on 3DRealCar-Part Dataset.} On the 3DRealCar-Part dataset, our SAV similarly outperforms all baseline methods, achieving 80.21/85.91 on mIoU/mAcc. Qualitative results are presented in Figure~\ref{fig:visual_results}. Similar results can also be found on the VehicleSeg10K Dataset.


\begin{figure}
    \centering
    \includegraphics[width=1\linewidth]{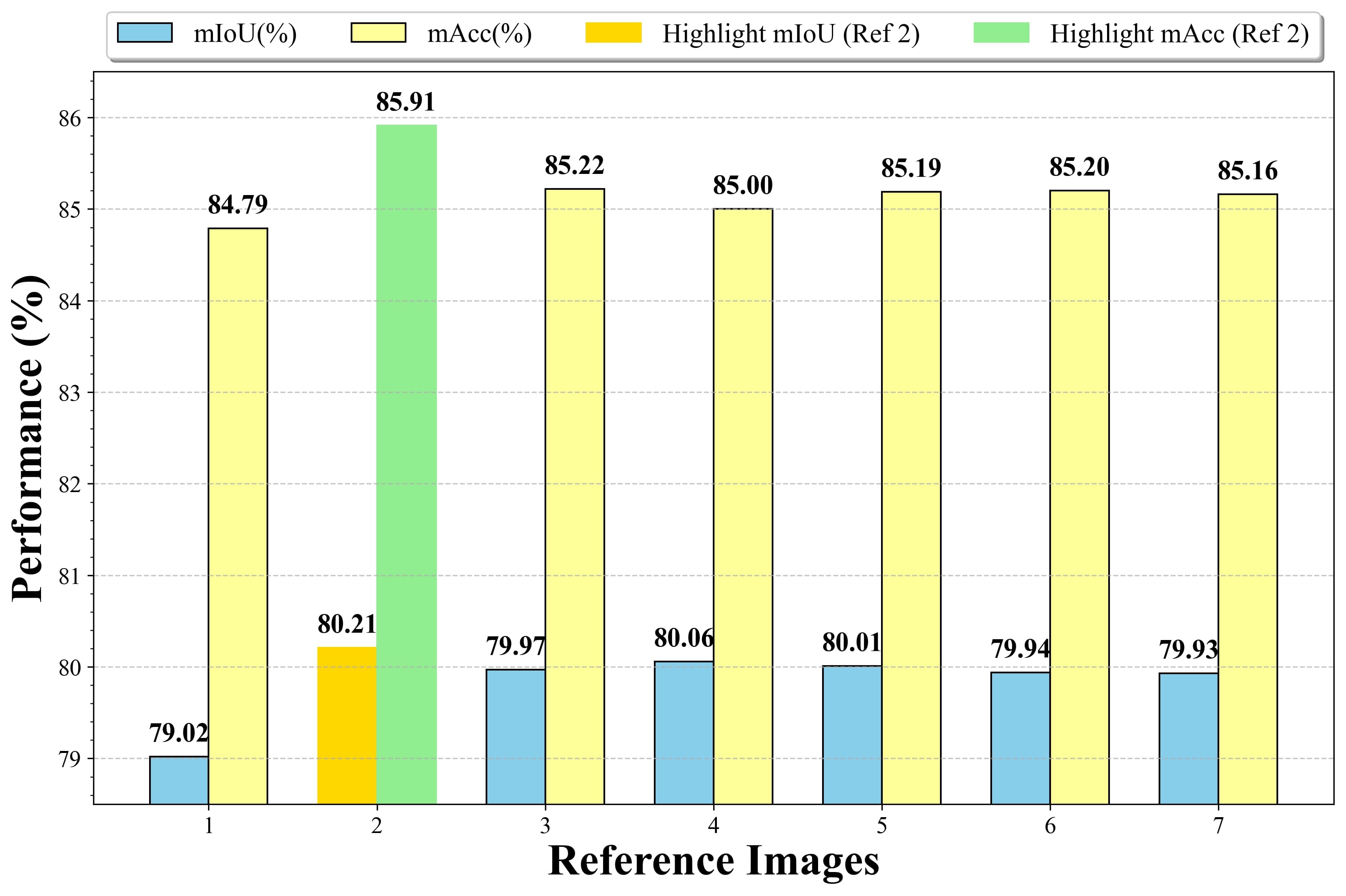}
    \caption{Impact of Reference Images on Dual Prototype Framework}
    \label{fig:reference_analysis}
\end{figure}

\begin{table}[htbp]
\centering
\caption{Performance Using Only Visual Prototypes}
\label{tab:reference_images_only}
\begin{tabular}{ccc}
\hline
\textbf{Reference Images} & \textbf{mIoU(\%)} & \textbf{mAcc (\%)} \\
\hline
1 & 57.03 & 61.25 \\
2 & 60.96 & 69.70 \\
\hline
\end{tabular}
\end{table}

\subsection{Component Analysis}

We use a modified SAM as the base model for component analysis. GTP denotes adding graph-enhanced textual prototypes to the base model, VP indicates incorporating visual prototypes through context sample retrieval, and RAM represents using the ROI attention mechanism for enhanced feature refinement. Table~\ref{tab:component_analysis} shows that the base model achieves 75.92\% mIoU and 61.22\% mAcc on the 3DRealCar-Part dataset. Adding GTP to integrate structural knowledge significantly boosts performance to 79.67\% mIoU and 65.12\% mAcc, representing a substantial +3.75\% mIoU improvement. When using VP alone with the base model, the framework achieves 77.43\% mIoU but remarkably high classification accuracy of 85.01\% mAcc, demonstrating the strong semantic discrimination capability of visual context samples. Finally, the complete framework combining all components (GTP + VP + RAM) achieves optimal performance with 80.21\% mIoU and 85.91\% mAcc. With each additional component, both segmentation and classification metrics steadily improve, indicating that each component contributes meaningfully to the model's performance. The notable improvements from GTP suggest that explicit structural modeling significantly enhances boundary delineation and spatial understanding, while VP excels at semantic classification through visual similarity matching. The addition of RAM provides targeted feature enhancement, particularly improving classification accuracy by selectively emphasizing discriminative characteristics within predicted regions.

\begin{table}[t]
\centering
\caption{Impact of Graph Attention Network (GATv2) Layer Depth on Performance}
\label{tab:gatv2_layer_analysis}
\begin{tabular}{c|cc}
\toprule
\textbf{GATv2 Layers} & \textbf{mIoU(\%)} & \textbf{mAcc(\%)} \\
\midrule
3 & 80.06 & 85.80 \\
\rowcolor[rgb]{0.95,0.95,0.95} 4 & \textbf{80.21} & \textbf{85.91} \\
5 & 79.90 & 85.46 \\
\bottomrule
\end{tabular}
\end{table}

\begin{table}[htbp]
\centering
\caption{Effect of Graph Enhancement on Textual Prototypes}
\label{tab:graph_enhancement}
\begin{tabular}{l|cc}
\hline
\textbf{Enhancement} & \textbf{mIoU(\%)} & \textbf{mAcc (\%)} \\
\hline
Without & 79.36 & 64.82 \\
With & 79.67 & 65.12 \\
\hline
\end{tabular}
\end{table}

\subsection{Ablation Study}

\begin{figure}
    \centering
    \includegraphics[width=\columnwidth]{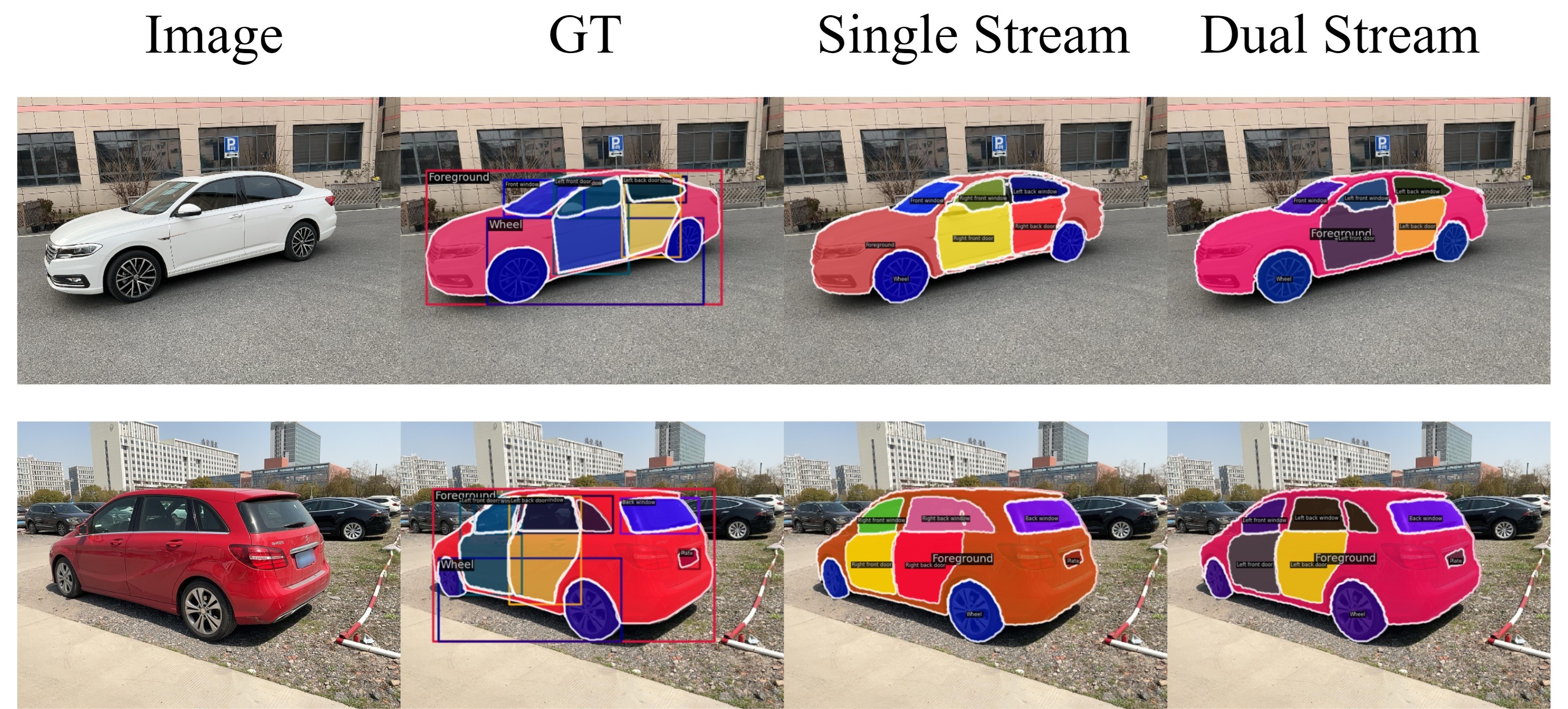}
    \caption{Visualization of our method in both single-stream and dual-stream scenarios on the 3DRealCar-Part Dataset.}
    \label{fig:dual}
\end{figure}

\begin{table}[t]
\centering
\caption{Ablation Study of Key Architectural Components}
\label{tab:component_analysis}
\begin{tabular}{cccc|cc}
\toprule
\textbf{Base} & \textbf{GTP} & \textbf{VP} & \textbf{RAM} & \textbf{mIoU (\%) } & \textbf{mAcc (\%)} \\
\midrule
\cmark & & & & 75.92 & 61.22 \\
\cmark & \cmark & & & 79.67 & 65.12 \\
\cmark & \cmark & & \cmark & 79.95 & 66.14 \\
\cmark & & \cmark & \cmark & 77.43 & 85.01 \\
\rowcolor[rgb]{0.95,0.95,0.95} \cmark & \cmark & \cmark & \cmark & \textbf{80.21} & \textbf{85.91} \\
\bottomrule
\end{tabular}
\end{table}

To validate the effectiveness of each component in our SAV framework, we conduct comprehensive ablation studies on the 3DRealCar-Part dataset. Each experiment is designed to isolate and evaluate the contribution of specific modules to the overall performance.

\noindent $\bullet$ \textbf{Effectiveness of Graph-Enhanced Textual Prototypes.}
To evaluate the contribution of our vehicle part knowledge graph, we compare model performance with and without graph enhancement for textual prototypes. As shown in Table~\ref{tab:graph_enhancement}, incorporating graph enhancement provides consistent improvements across metrics. The model achieves 79.67\% mIoU with graph enhancement compared to 79.36\% without it. While the improvement appears modest, this consistent gain across metrics indicates that explicit structural modeling of vehicle part relationships provides valuable complementary information for spatial adjacency understanding that pure textual embeddings cannot capture.

\noindent $\bullet$ \textbf{Impact of Visual Context Sample Augmentation.}
We systematically analyze the effect of the number of reference images used in our context sample augmentation mechanism. The experiments vary the number of retrieved reference images from one to seven while maintaining consistent retrieval quality. As illustrated in Figure~\ref{fig:reference_analysis}, performance initially improves when increasing reference images from one to two (mIoU: 79.02\% → 80.21\%), achieving the optimal configuration. However, further increasing to three or more reference images leads to performance degradation. This performance degradation occurs because excessive reference images introduce conflicting visual information that hampers the model's ability to focus on relevant contextual features. The optimal two-image configuration provides sufficient visual diversity.

\noindent $\bullet$ \textbf{Analysis of Graph Attention Network Depth.}
We investigate the optimal depth for our Graph Attention Network (GATv2) in the vehicle part knowledge graph module by varying the number of layers from 3 to 5. Table~\ref{tab:gatv2_layer_analysis} demonstrates that 4 layers provide the optimal balance, achieving 80.21\% mIoU and 85.91\% mAcc. The 3-layer configuration yields 80.06\% mIoU and 85.80\% mAcc, indicating insufficient depth to fully capture complex structural relationships through multi-hop attention propagation. Conversely, the 5-layer configuration shows decreased performance (79.90\% mIoU, 85.46\% mAcc), suggesting over-smoothing of node features where deeper layers dilute part-specific discriminative information. The 4-layer architecture effectively balances the trade-off between capturing long-range structural dependencies and preserving local part characteristics.


\noindent $\bullet$ \textbf{Comparison between Visual-Only and Full Framework Approaches.}
To demonstrate the necessity of our dual-stream approach, we evaluate model performance using only visual prototypes without structural guidance. As shown in Figure~\ref{fig:dual}, our dual-stream framework consistently outperforms single-stream configurations across diverse vehicle types and viewpoints. Table~\ref{tab:reference_images_only} shows that while increasing reference images from one to two improves visual-only performance (57.03\% to 60.96\% mIoU), this approach significantly underperforms compared to our complete framework, validating that structural knowledge and visual context provide complementary rather than redundant information, with structural constraints being essential for achieving high-quality segmentation boundaries.

\noindent $\bullet$ \textbf{Model Parameter and Efficiency Analysis.}
We analyze the computational efficiency and parameter requirements of our approach compared to existing methods. As shown in Table~\ref{tab:benchmark_comparison}, while our model has 313.2M parameters (larger than lightweight architectures), the substantial performance gains (80.21\% mIoU vs. 74.97\% for the next best method) justify this increased complexity. The parameter efficiency remains competitive with other large models while achieving significantly superior performance in vehicle part segmentation tasks.

\subsection{Limitation Analysis}
Despite promising results, several limitations merit consideration and suggest directions for further research. Our model is constrained by the inherent limitations of the SAM architecture. Although it has been successfully adapted for zero-prompt operation and multi-class segmentation, the decoder still prioritizes object-centric segmentation rather than part-level decomposition. This occasionally results in segmentation errors at the boundaries between adjacent parts with similar visual properties. Computational complexity also poses challenges for real-time applications. Our dual prototype framework involves multiple processing streams, which add overhead compared to simpler models. Efficiency optimization without compromising performance remains an important research direction.

\section{Conclusion and Future Works} \label{sec:conclusion}
In this paper, we introduced SAV, a novel framework that transforms SAM for part-level vehicle segmentation without explicit prompts. By incorporating a partial semantic graph and contextual priors, our approach achieves state-of-the-art performance on multiple benchmarks while maintaining efficient operation. The graph-enhanced textual prototypes capture structural relationships between vehicle parts, while the context-aware visual prototypes provide complementary fine-grained details. Together, these components enable precise and semantically consistent segmentation across diverse vehicle types and viewing conditions.

In future work, we plan to explore more efficient architectural designs that maintain the benefits of our dual prototype approach while reducing computational requirements. Additionally, we aim to extend our approach to video segmentation of vehicle parts, where temporal consistency provides additional constraints that could further improve segmentation quality.

\small{ 
\bibliographystyle{IEEEtran}
\bibliography{reference}
}

\end{document}